\documentclass[final,1p,times]{elsarticle}

\usepackage{amssymb}

\usepackage{graphicx}
\usepackage{amsmath}
\usepackage{bm}
\usepackage{amssymb}
\usepackage{booktabs}
\usepackage{float}
\usepackage{multirow}
\usepackage{tabularx}
\usepackage{array}
\usepackage{tabulary}
\usepackage{colortbl}
\usepackage{soul}
\usepackage{color}
\usepackage{amsmath}
\usepackage{url}
\usepackage[ruled,vlined]{algorithm2e}

\newcommand{\imgStubPDFpage}[4]{\begin{minipage}{#1\textwidth}\begin{center}
 \includegraphics[#2]{#3}\\
#4\end{center}\end{minipage}}

\definecolor{cadmiumyellow}{rgb}{1.0, 0.96, 0.0}
\definecolor{amber}{rgb}{1.0, 0.75, 0.0}
\definecolor{mikadoyellow}{rgb}{1.0, 0.77, 0.05}

\newcommand{\Stress}[1]{\textbf{\textit{#1}}}
\newcommand{\Yellow}[1]{\textcolor{mikadoyellow}{#1}}
\newcommand{\Red}[1]{\textcolor{red}{#1}}

\begin{document}

\begin{frontmatter}



\title{Reconstruct Face from Features Using GAN Generator as a Distribution Constraint\tnoteref{label1}} 


\author[inst1]{Xingbo Dong}
\author[inst2]{Zhihui Miao}
\author[inst3]{Lan Ma}
\author[inst3]{Jiajun Shen}
\author[inst4]{Zhe Jin}
\author[inst5]{Zhenhua Guo}
\author[inst1]{Andrew Beng Jin Teoh\corref{cor1}}

\cortext[cor1]{Corresponding author: Andrew Beng Jin Teoh (e-mail: bjteoh@yonsei.ac.kr).}
\tnotetext[label1]{This work was supported the Brain Pool Program through the National Research Foundation of Korea (NRF) funded by the Ministry of Science and ICT under Grant 2021H1D3A2A01099396.}
\affiliation[inst1]{organization={Yonsei University},
 city={Seoul},
 country={Republic of Korea}}

\affiliation[inst2]{organization={Fuzhou University},
 city={Fuzhou},
 country={China}}

\affiliation[inst3]{organization={TCL AI Lab},
 city={Shenzhen},
 country={China}}

\affiliation[inst4]{organization={Anhui University},
 city={Hefei},
 country={China}}
 
\affiliation[inst5]{organization={Tsinghua Shenzhen International Graduate School},
 city={Shenzhen},
 country={China}}
 
\begin{abstract}
Face recognition based on the deep convolutional neural networks (CNN) shows superior accuracy performance attributed to the high discriminative features extracted. 
Yet, the security and privacy of the extracted features from deep learning models (deep features) have been often overlooked. This paper proposes the reconstruction of face images from deep features without accessing the CNN network configurations as a constrained optimization problem. Such optimization minimizes the distance between the features extracted from the original face image and the reconstructed face image. Instead of directly solving the optimization problem in the image space, we innovatively reformulate the problem by looking for a latent vector of a GAN generator, then use it to generate the face image. The GAN generator serves as a dual role in this novel framework, i.e., face distribution constraint of the optimization goal and a face generator. 
On top of the novel optimization task, we also propose an attack pipeline to impersonate the target user based on the generated face image. Our results show that the generated face images can achieve a state-of-the-art successful attack rate of 98.0\% on LFW under type-I attack @ FAR of 0.1\%. Our work sheds light on the biometric deployment to meet the privacy-preserving and security policies.

\end{abstract}





\end{frontmatter}


\section{Introduction}
\label{sec:introduction}
With the advancement of deep learning, the face has become one of the most popular biometric traits because of its convenience and superior recognition performance. However, while the deployment of face recognition (FR) facilitates the comfort and efficiency of identity management, it also evokes inevitable challenges to privacy and security because of the permanent linkage of the face data and the corresponding owner's identity. One main concern of the face recognition system is that the disclosure of face image may expose the private and sensitive information of the user, and such information can be exploited to gain illegal access to the user's system \cite{patel2015cancelable,gomez2020reversing}. Face database breaches are not uncommon and they have been reported in the news media. For example, in 2019, 2.56 million people's personal information, including ID cards and photos, were leaked from an AI company \footnote{https://www.biometricupdate.com/201902/facial-recognition-company-sensenets-silent-following-data-leak}. In 2020, Clearview AI, a company that owns over 3 billion photos in the database, suffered a data breach \footnote{https://www.forbes.com/sites/kateoflahertyuk/2020/02/26/clearview-ai-the-company-whose-database-has-amassed-3-billion-photos-hacked}.

On the other hand, generative adversarial networks (GANs) have actively been utilized to manipulate face images or videos for malicious purposes such as face morphing attacks, face swapping (Deepfake) etc.  \cite{deepfakes, isola2017image,korshunova2017fast,nirkin2018face}. Such attacks require the knowledge of the target user's face image. On the contrary, fewer investigations have been done to reconstruct face images from the face features extracted by the deep model \Stress{without} knowing the target user's face image. 

\begin{figure}
\centering
\imgStubPDFpage{0.99}{page=5,width=0.99\linewidth,trim=0cm 10.8cm 0cm 0cm, clip}{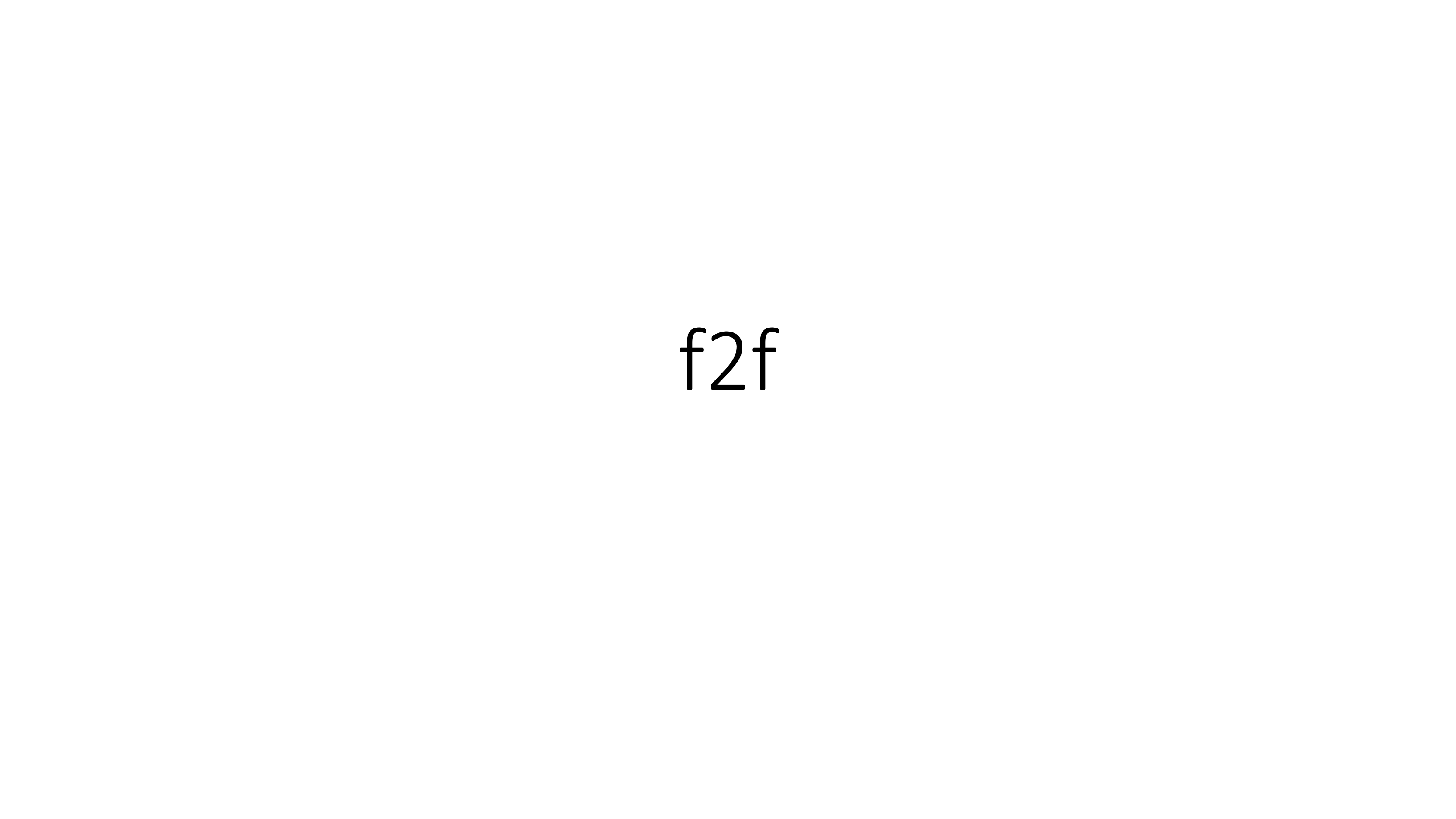}{} 
\caption{We propose a novel pipeline to reconstruct high visual quality face images from their corresponding deep features by using the GAN generator as a distribution constraint and considering the feature extractor as a black box. This figure demonstrates the reconstructed face image on the colorFERET dataset (type-I); \Yellow{Yellow} numbers are the normalized similarity scores between features of reconstructed and bona fide faces, and they are above the system's decision threshold (0.6324) at FAR=0. \Red{Red} numbers are COTS confidences (\%) of being the same person compared with bona fide. 
\label{Figurehomepage}}
\end{figure}

Only a few works have previously engaged in the face images reconstruction from the face deep features\footnote{We denote features extracted from deep learning models as deep features or deep templates. They are used interchangeably.}. Previous works have typically been based on convolutional networks \cite{cole2017synthesizing} or de-convolutional neural networks \cite{mai2018reconstruction}. Among them, \cite{mai2018reconstruction} is specifically proposed to reconstruct the face image from its deep feature. The generated face images by \cite{mai2018reconstruction} achieve a state-of-the-art attack success rate. However, the generated face images in \cite{cole2017synthesizing,mai2018reconstruction} are in low-resolution, often leaving the synthetical artifacts that can let human or machines spot easily (see Figure \ref{Figure:f2f_reconstruct}); furthermore, a GAN model is also required in \cite{mai2018reconstruction} to generate a tremendous amount of images for the network training. In \cite{dong2021towards}, a neural network is established to learn a mapping between the latent vector space of the StyleGAN2 \cite{Karras2019StyleGAN2} and the feature vector space of the face feature extractor. Given a feature vector, the proposed model can predict the corresponding latent vector, and high-quality face images (1024$\times$1024) can be generated subsequently. However, the mapping method in \cite{dong2021towards} leads to the artifacts in the reconstructed face and can only achieve an attack success rate of 10.1\%@1\%FAR (false accept rate) when compared with bona fide (genuine) face images on the LFW dataset.

\begin{figure*}[t!]
 \centering
 \includegraphics[width=0.99\linewidth]{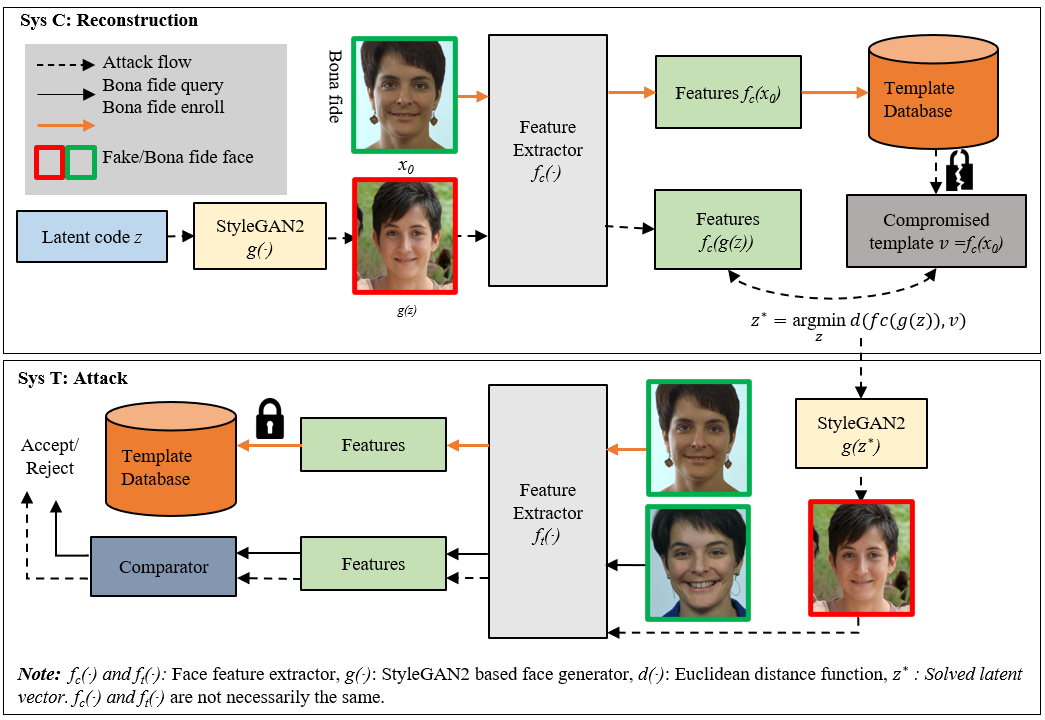}
 \caption{Overview of the framework for reconstructing face images from the corresponding compromised deep templates. The constrained optimization problem aims at solving for a latent vector in the GAN generator's latent space, and then the latent vector is used to generate the face image. To simulate the impersonate attack, the reconstructed face images from Sys C are used to attack Sys T. The proportion of attack attempts that are falsely accepted by Sys T is evaluated as the attack performance. Sys C does not necessarily share the same feature extractor with Sys T. Best viewed in color.}
 \label{Figure:f2f_Overview}
\end{figure*}

Our work focuses on reconstructing face images based on deep features. We demonstrate the effectiveness of our approach by launching impersonate attacks on the face recognition system using the generated face images. 
More specifically, given $x$ is the enrolled face image, and $ f(\cdot) $ is the face feature extractor, we assume that the stored template $ \bm{v}=f(\bm{x}) \in \mathcal{V} $ is known to the adversary, where $ \mathcal{V} $ denotes the feature space and $x$ is unknown to the adversary. We also assume the deep feature extractor is a black box wherein the details of the extractor are not known to the adversary. Still, the adversary can generate unlimited input-output data pairs from the deep feature extractor. To achieve the goal mentioned above, we reformulate the face image reconstruction problem as a constrained optimization problem and explore a solution based on the Genetic Algorithm (GA). As shown in Figure \ref{Figurehomepage}, our method can generate high-quality images from the deep features with semantic details. Furthermore, the generated face can be utilized to access a target FR system based on the impersonate attack. The overview of the proposed pipeline is shown in Figure \ref{Figure:f2f_Overview}.

In summary, this paper makes the following contributions:
\begin{enumerate}
\item We propose a novel framework for reconstructing face images from deep features extracted by deep learning models. We first reformulate the face image reconstruction problem as a constrained optimization problem. Then, a pipeline of face image reconstruction is proposed by searching for a latent vector in the GAN generator's latent space, which is used to generate the face image. The GAN generator serves as a face distribution constraint of the optimization goal and a face generator. Subsequently, we adopt a Genetic Algorithm (GA) to solve the optimization problem and validate the feasibility. 
\item We evaluate our reconstructed face by launching an impersonate attack to access a target face recognition system with the same or different feature extractor. Our method can achieve state-of-the-art attack performance on the LFW datasets (98.0\%@0.1\%FAR) and comparable performance on colorFERET datasets (76.0\%@0.1\%FAR).
\item The generated face images are also evaluated with three Commercial-Off-The-Shelf (COTS) face liveness detectors. The result shows that the generated high resolution face images can fool the COTS detectors with a pass rate of 88.6\%, 62\%, and 98\%. 
\end{enumerate}

\section{Related Work}
With the rapid developments in computing hardware, big data, and novel algorithms, deep learning-based FR techniques have fostered numerous startups with practical applications in the past five years. Massive deployment of FR systems based on the deep learning models \cite{schroff2015facenet,deng2018arcface,Parkhi15,ouyang2015deepid,liu2017sphereface,wen2016discriminative,zhang2017range,wang2018cosface} draws the public's attention to the privacy and security concerns of reconstructing face images from deep features \cite{ranjan2018deep,wang2020deep}. According to the techniques used, face image reconstruction can be divided into conventional and deep learning methods. Several conventional face image reconstructions are based on the optimization algorithms, such as hill-climbing \cite{adler2003sample,feng2014masquerade} and radial basis function (RBF) regression \cite{mignon2013reconstructing}. 
On the other hand, given the adversary's knowledge of the extractor, face image reconstruction can be classified into two main branches, i.e., white-box-based (wherein feature extractor is known) and black-box-based (wherein feature extractor is unknown) approaches. 
This section mainly discusses the deep learning-based methods as we focus on deep templates. 

A method to invert FaceNet face embedding \cite{schroff2015facenet} to face images is presented in \cite{zhmoginov2016inverting}. The inverting task is formulated as a minimization problem that minimizes the template difference between original and reconstructed images. To achieve this task, a regularization function based on the intermediate layer of the feature extractor is utilized. However, the detailed parameter of the extractor may not be available in reality. Later, a method to generate normalized face images is proposed \cite{cole2017synthesizing}. Although the motivation \cite{cole2017synthesizing} is different from reconstructing a face from features, the normalized face images are generated from the FaceNet features based on differentiable image warping by combining landmark and texture information. However, the landmark and texture information is required, and the last convolution layer of the extractor is also utilized in the pipeline, which may be impractical in reality for face image reconstruction.

A de-convolutional neural network is used in \cite{mai2018reconstruction} to achieve the face image reconstruction task because of its up-sampling capability. In \cite{mai2018reconstruction}, the feature extractor is regarded as a black box that the adversary does not necessarily know. A cascade of multiple stacked de-convolution blocks and a convolution block, namely neighborly de-convolutional neural network (NbNet), is designed to generate the output face images. GAN synthesizes face images, and two benchmark face datasets are utilized for training the NbNet. The experiment results indicate that 95.20\% of the generated face images can successfully access the face recognition system enrolled with the same face image at an FAR of 0.1\% (type-I attack in \cite{mai2018reconstruction}). Later in \cite{keller2021inverting}, by utilizing the NbNet, the authors propose to reconstruct the face image from a binary template produced by a given binarization method. Other methods, such as iterative sampling of random Gaussian blobs \cite{razzhigaev2020black} and autoencoder \cite{ahmad2020resist}, have also been reported in the literature. However, those methods can only produce low-resolution and vision-unfriendly images which the naked eye can easily spot. 

In \cite{dong2021towards}, a framework to reconstruct high-quality face images from deep features is proposed. A neural network is established to learn a mapping between the latent vector space of the StyleGAN2 \cite{Karras2019StyleGAN2} and the feature vector space of the face feature extraction model. Given a feature vector, the model predicts the corresponding latent vector, and face images can be generated subsequently. However, the work based on the vanilla fully connected neural network \cite{dong2021towards} is suboptimal for this task. The stochastic gradient descent algorithm can be trapped in a local minimum. Hence, the method can only achieve a successful attack rate of 10\% on LFW under type-I attack @1\%FAR. As a result, most reconstructed face images show poor visual similarity compared with the Bona fide face images (see Figure \ref{Figure:f2f_reconstruct_hq}).


One shortcoming of the methods mentioned above is that only low-resolution face images can be generated. 
For instance, the output image size in \cite{mai2018reconstruction} is $160\times$160, and the size in \cite{cole2017synthesizing,zhmoginov2016inverting} is $224\times$224. 224 may be enough for face recognition and liveness detection, but higher resolution is usually preferred and considered a prerequisite for face recognition \cite{lu2018deep,li2019low,jourabloo2018face}. On the other hand, although face image reconstruction has been achieved in \cite{dong2021towards}, tremendous artifacts exist, and the attack success rate still leaves a large room for improvement (see \textbf{section 4.2}). To address the discussed problems, we reformulate the face reconstruction task as a novel constrained optimization problem and solve it by GA due to its population-based evolution property; hence, highly visually similar face images reconstructed from their corresponding features can be achieved.

\section{Face Reconstruction on Generator-based Distribution Constraint}
In this work, we assume that the stored template $\bm{v}=f(\bm{x}_0) \in \mathcal{V}$ is known to the adversary, where $\mathcal{V}$ denotes the feature space and $\bm{x}_0$ is the registered face images. We highlight that big quantities of templates can be compromised and available to the adversarial systems, but we focus on \textbf{single compromised template} case. We also assume that the adversary can exploit the deep feature extractor \Stress{$f(\cdot)$} to generate unlimited input-output data pairs. However, the deep feature extractor is regarded as a black box that is not necessarily known to the adversary. Finally, it is worth pointing out that \textit{we assume that images of the target person are not used to train the target face recognition system}. 

The goal of the adversary is to reconstruct a \Stress{visually similar} face image $\bm{x}^*$ from the compromised template $\bm{v}$; meanwhile the features extracted from $\bm{x}^*$ should be \Stress{closer} to the original feature $\bm{v}$. The goal is formulated as: 
\begin{equation}
\begin{split}
\bm{x}^* & = \operatorname*{arg\,min}_{\bm{x}} d(f(\bm{x}),\bm{v}), \\ & 
s.t. {~\bm{x}\in \bm{X}}, \label{eq.initalgoal}
\end{split}
\end{equation}
where $d(\cdot)$ is a distance function (e.g., Euclidean distance), $f(\cdot)$ indicates a deep learning-based feature extraction function and $\bm{v} = f(\bm{x}_0)$ is the compromised template in the database, and $\bm{X}$ denotes the domain of human face. Note that other feature extractors such as LBP \cite{deng2018compressive,ahonen2006face} and HOG \cite{dalal2005histograms}, can also be used, as we treat it as a black box.

\begin{algorithm}[t!]
\caption{Solving the latent vector by GA.\label{alg.ga}}
\SetAlgoLined
\begin{flushleft}
\textbf{INPUT:} Compromised template $v$\\
Face generator $g$\\
Face encoder $f$\\
GA population size $t$\\
GA selection rate $s$\\
\textbf{OUTPUT:} Solved latent vector $ \bm{z}^* $
\end{flushleft}

Randomly initiate GA population as $Z=\{z_i| i = 1,2,...,t\}$

\Repeat{converged}{
 Step1: Generate face images $X = \{g(z_i)| i = 1,2,...,t\}$\;
 Step2: Generate deep features $V = \{f(g(z_i))| i = 1,2,...,t\}$\;
 Step3: Compute the distance $d(f(g(\bm{z})),\bm{v})$ as the fitness value for each $z$\;
 Step4: Top $s\times t$ best $z$ are selected as the parents $Z_{parents}$ of next generation\;
 Step5: Generate new child $Z_{children}$ by crossover and mutation operations on the $Z_{parents}$\;
 Step6: Generate the next generation population:$Z=\{Z_{parents},Z_{children}\}$ \;
 }
 Step7: return the best individual of the last generation population as $\bm{z}^*$.
\end{algorithm}
One challenge for the above optimization goal is how to impose a constraint to the $\bm{x}$, i.e., constraining the $\bm{x}_0$ to be a human face. One naive way is representing the constraint in terms of probability distribution:
\begin{equation}
\begin{split}
\bm{x}^* & = \operatorname*{arg\,min}_{\bm{x}} d(f(\bm{x}),\bm{v}),\\ & s.t. {~\bm{x}\sim P_{face}(x)}, \label{eq.goal2}
\end{split}
\end{equation}
where $P_{face}(x)$ denotes the distribution of face images. However, as $P_{face}(x)$ cannot be explicitly defined, it remains challenging to incorporate $P_{face}(x)$ directly into the optimization objective function when solving the above problem. 
Motivated by GAN's distribution modeling capability\cite{goodfellow2014generative,creswell2018generative,li2017alice,mao2017least,goodfellow2016nips}, we propose to relax the above optimization function by transferring the optimization solution space from the image space to the GAN's latent space. Generative models, such as StyleGAN \cite{karras2019style,Karras2019StyleGAN2}, learn a mapping relationship between the prior distribution $\pi(z)$ and the training set's pixel probability distribution $p(x)$. The prior distribution $\pi(z)$ is usually a Gaussian distribution, and the generator's output follows the distribution of face images.
Since StyleGAN is powerful for face generation, we opt for StyleGAN2 \cite{Karras2019StyleGAN2} generator as the constraint of the optimization goal. Hence Eq.(\ref{eq.goal2}) can be reformulated as:
\begin{equation}
\begin{split}
\bm{z}^* & = \operatorname*{arg\,min}_{\bm{z}} d(f(g(\bm{z};\theta)),\bm{v}),\\
s.t. & {~\bm{z}\sim \mathcal{N}(0,I)}, \\
&{P_{g(\bm{z};\theta)} \approx P_{face}(x)}, \label{eq.finalgoal}
\end{split}
\end{equation}
where $\theta$ is the parameter of StyleGAN2, $g(\cdot)$ indicates a pre-trained StyleGAN2 based face generation model function. The constraint $\bm{x}\sim P_{face}(x)$ in Eq. (\ref{eq.goal2}) is eliminated at the cost of a strong assumption $P_{g(\bm{z};\theta)} \approx P_{face}(x)$; in another word, we assume that the distribution of well-trained generator's output image $P_{g(\bm{z};\theta)}$ resembles real human face distribution $P_{face}(x)$, i.e., $P_{g(\bm{z};\theta)} \approx P_{face}(x)$. The optimization goal is now looking for the best latent vector $z$, which is more easily operated. 


As opposed to \cite{mai2018reconstruction} and \cite{dong2021towards}, we aim at reconstructing high visual quality and similarity with the bona fide face images. We propose a novel way to achieve template inverting tasks of Eq. (\ref{eq.finalgoal}) by incorporating a standard Genetic Algorithm (GA)\cite{srinivas1994genetic}. The detailed processing pipeline is shown in Algorithm \ref{alg.ga}.

As the investigation \cite{shen2020interpreting} suggests, the latent code of well-trained generative models learns a disentangled representation. Given the GA's ability to search for the best solution with a vast population of candidates representing the various face attribute combinations pulled from the latent vector, GA is preferable for the above-formulated problem.

\section{Experiments and Results}
\label{sec:guidelines}

\subsection{Settings}
In our experiment, a pre-trained StyleGAN2 model\footnote{https://nvlabs-fi-cdn.nvidia.com/StyleGAN2/networks/StyleGAN2-ffhq-config-f.pkl} is adopted. The pre-trained model is trained on the FFHQ dataset \cite{karras2019style}, consisting of 70,000 high-quality images at 1024×1024 resolution. We trained three InsightFace models with different backbones, including ResNet-50, Xception, and InceptionResNet on MS-celeb-1m dataset \cite{guo2016ms} in our experiments.

To evaluate the performance of the proposed method, the Labeled Faces in the Wild (LFW) \cite{huang-2008-LFW,LFWTechUpdate}, and color FERET (colorFERET) \cite{phillips1998feret,phillips2000feret} datasets are adopted in our experiments. LFW and colorFERET are selected as they represent unconstrained or semi-constrained environments: 1) face images in LFW are collected from the Internet and are usually captured in unconstrained environments; 2) face images in the color FERET database are captured in a semi-controlled environment with 13 different poses. 

To initiate the GA, the default population size was set to 256 individuals, and the selection rate was set at 0.3; while the mutation ratio was set at 0.3, the cross-over point is selected randomly. The mutation is performed by adding a random number drawn from a standard Gaussian distribution to each entry of the parent vector. The algorithm ends if the fitness value does not decrease for the last 20 generations or the algorithm has already reached 1000 generations. 

To simulate the attack risks, in reality, two simulation scenarios \cite{dong2019security} are adopted as below (also see Figure \ref{Figure:f2f_Overview}):

\Stress{Compromised system ($Sys~C$)}: This is a biometric system compromised by the adversary. In this subsystem, we assume that 1) the stored template $\bm{v}=f_c(\bm{x}_0)$ of the target user is compromised by the adversary; 2) the deep feature extractor can be utilized by the adversary to generate unlimited input-output data pairs; 3) the deep feature extractor is a black-box to the adversary in that the adversary does not necessarily know the parameters. InsightFace model with ResNet-50 backbone is adopted as the feature extractor in $Sys~C$.

\Stress{Targeted system ($Sys~T$)}: This is a biometric system subject to a reconstruction attack. Each user's reconstructed face image $\bm{\hat{x}}$ from a template stored in the $Sys~C$ was generated and used to break into $Sys~T$ as shown in Figure 2. The feature extractor $f_f(\cdot)$ used in the $Sys~T$ could be different from $Sys~C$, and the user's enrolled image could also be different from $Sys~C$. We consider both cases in our subsequent experiment. Three InsightFace models with ResNet-50, InceptionResNet, and Xception backbone are adopted as the candidate feature extractors in $Sys~T$.

\begin{figure*}[t]
 \centering
 \includegraphics[width=0.99\linewidth]{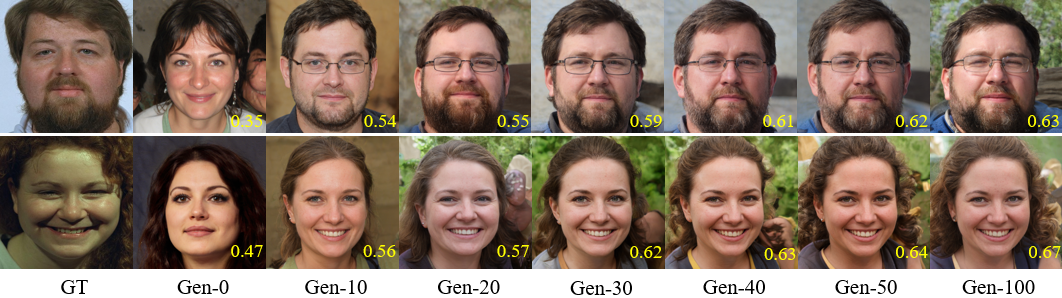}
 \caption{Visual demonstration of the evolution of the GA. GT refers to the ground truth face images, and images with normalized similarity score (\Yellow{yellow} number) are the best candidate images (type-I) selected at the $i$-th generation (Gen-$i$) of the GA.\label{Figure:f2f_Evolution}}
\end{figure*}

Each user's face image is reconstructed in the attack simulation based on one compromised template in $Sys~C$. Then, three GA threads are initiated for each user simultaneously, and the best individual candidate among the three threads is selected to reconstruct the final face image. Subsequently, in $Sys~T$, the face template is extracted from the reconstructed face image. Finally, three types of similarity scores are calculated: 1) \Stress{mated-attack score} is computed by comparing reconstructed face features with the real \Stress{mated} face of the same person; 2) \Stress{genuine scores} are computed as the similarity between bona fide features from the same person; and 3) \Stress{imposter scores} are computed as the similarity between bona fide features from different persons. 

The successful attack rate (SAR) is calculated as the proportion of mated-attack attempts that are falsely declared to match the template of the same user at the given similarity threshold under different FAR, i.e., the ratio of mated-attack scores above the similarity threshold. The threshold is determined based on the normal genuine and imposter scores. SAR can be regarded as the false match rate under attack, and a higher SAR implies the higher performance of the reconstruction attack and vice versa. Similar to \cite{mai2018reconstruction}, when the real template is from the same compromised face image in $Sys~C$, the attack is referred to as a type-I attack. When the real template is from the different face images of the same person, the attack is regarded as a type-II attack.

\begin{figure*}[]
 \centering
 \includegraphics[width=0.99\linewidth, trim=0.8cm 8.8cm 2cm 0cm, clip]{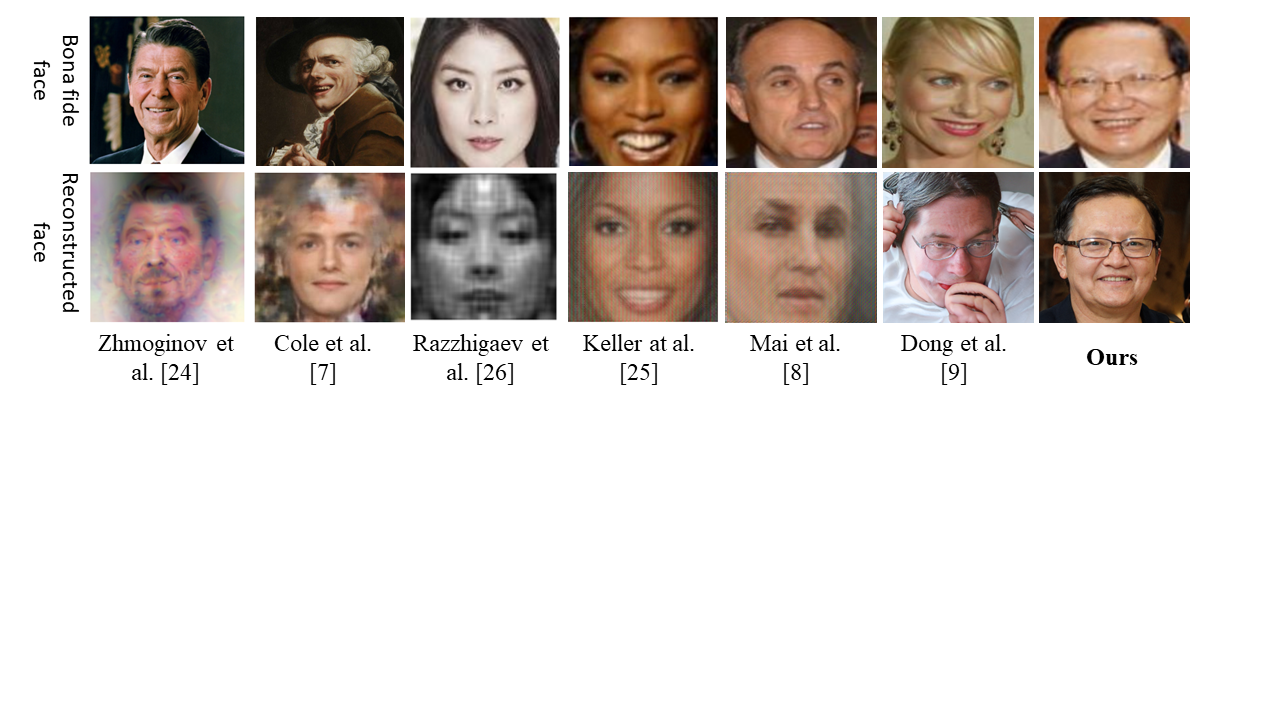}
 \caption{Comparison of visual quality with existing works.\label{Figure:f2f_reconstruct_hq}}
\end{figure*}

\begin{figure*}[]
 \centering
 \includegraphics[width=0.99\linewidth]{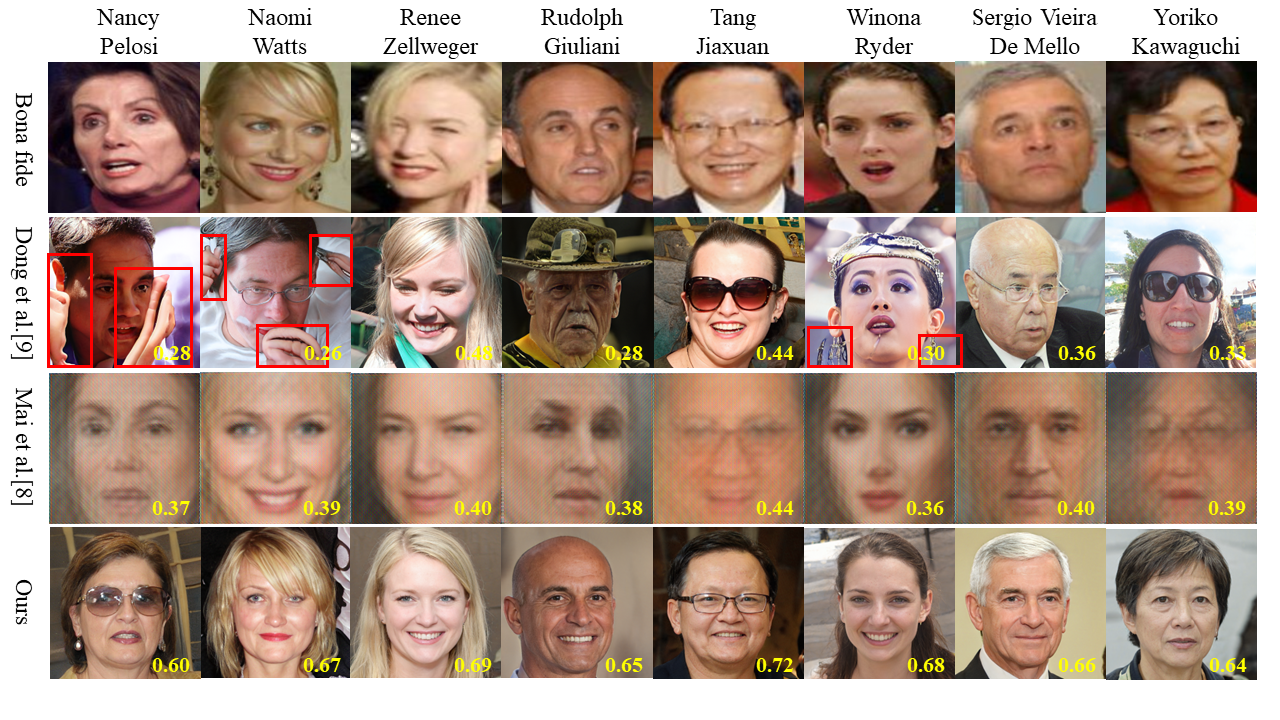}
 \caption{Visual demonstration of reconstructed face images from the LFW dataset (type-I). \Yellow{Yellow} numbers are the normalized similarity scores between features of the corresponding image and bona fide face, they are above the FAR=0 decision threshold (0.5212), and red boxes indicate the artifacts caused by sub-optimal latent codes produced by \cite{dong2021towards}.\label{Figure:f2f_reconstruct}}
\end{figure*}
\subsection{Qualitative Evaluation}
We first examine the effectiveness of the genetic algorithm. Examples of the generated face images from the $0^{th}$ to $100^{th}$ generation of GA are shown in Figure \ref{Figure:f2f_Evolution}. We can find that although images in the initial generation show distinct faces, as the GA manages to search for an optimal latent vector, the algorithm can generate faces visually similar to the corresponding bona fide images as the generation evolves. As Figure \ref{Figure:f2f_Evolution} suggests, 20 generations of evolution are enough to produce visually similar fake face images. Furthermore, the similarity between the target and face templates increases as the generation evolves. For example, as Figure \ref{Figure:f2f_Evolution} shows, the fitness value can increase to 0.63 from 0.35 for the first user in the first row. This indicates that the proposed framework is effective and efficient. 

Next, a comparison of image visual quality with existing works is shown in Figure \ref{Figure:f2f_reconstruct_hq}. Compared with \cite{zhmoginov2016inverting,cole2017synthesizing,razzhigaev2020black,keller2021inverting,mai2018reconstruction}, our generated face images have higher visual quality and higher resolution. Though high resolution images can be generated by \cite{dong2021towards}, but many artifacts are founded. 

Examples of bona-fide images and their generated face images based on randomly selected subjects in colorFERET and LFW are shown in Figure \ref{Figurehomepage} and Figure \ref{Figure:f2f_reconstruct}. The final succeeded fake face instances show high visual similarity compared with the corresponding bona fide images on both LFW and colorFERET datasets. The visual results indicate that the proposed method can generate high-quality and similar face images from the corresponding face features with StyleGAN's generator. Additionally, compared with \cite{mai2018reconstruction}, our generated face images have higher visual quality and higher resolution. However, many artifacts are founded on \cite{dong2021towards} because of sub-optimal latent codes produced by \cite{dong2021towards};additionally, better visual similarity can be observed in the images generated by our proposed novel pipeline. 


We further evaluate the generated images from colorFERET with one COTS\footnote{https://www.faceplusplus.com.cn/face-comparing/} face comparator, and the confidence that the images belong to the same person is also shown in Figure \ref{Figurehomepage}. The results suggest that our reconstructed high quality face images can fool even COTS.

\subsection{Quantitative Evaluation}
We also quantitatively evaluate the performance of the proposed reconstruction scheme based on type-I and type-II attacks with the same and different feature extractors. We first compute the true accept rate (TAR) at FAR on the deep features under the normal situation in $Sys~T$. Specifically, in the LFW dataset, 50 users with at least 10 images are selected randomly to compute the FAR and TAR, while in colorFERET, 50 users with at least two images are selected randomly. Then, in the reconstruction process, 50 users from LFW and 50 from colorFERET are chosen randomly to perform the face image reconstruction attack. The SARs of type I and II attacks are evaluated subsequently, and the results are shown in Table \ref{tab.syst_res50}.

\begin{table}[H]
\centering
\caption{Attack performance in Sys T with a insightFace model on various backbones (Threshold is normalized between 0 to 1).\label{tab.syst_res50}}
\begin{tabular}{c|c|c|c|c|c} 
\hline
Dataset & FAR(\%) &{\begin{tabular}[c]{c}Standard\\TAR(\%)\end{tabular}}& Threshold & {\begin{tabular}[c]{c}Type-I\\SAR(\%)\end{tabular}}& {\begin{tabular}[c]{c}Type-II\\SAR(\%)\end{tabular}} \\ 
\hline
\multicolumn{6}{c}{{\cellcolor[rgb]{0.773,0.773,0.773}}Resnet-50 backbone (same extractor)} \\ 
\hline
\multirow{4}{*}{LFW} & 0.00 & 26.03 & 0.5212 & 86.00 & 13.63 \\ 
\cline{2-6}
 & 0.10 & 50.28 & 0.4734 & 98.00 & 29.37 \\ 
\cline{2-6}
 & 1.00 & 73.61 & 0.4300 & 100.00 & 50.92 \\ 
\cline{2-6}
 & 10.00 & 92.66 & 0.3780 & 100.00 & 76.06 \\ 
\hline\hline
\multirow{4}{*}{\begin{tabular}[c]{@{}c@{}}color\\FERET\end{tabular}} & 0.00 & 58.66 & 0.6324 & 38.00 & 13.44 \\ 
\cline{2-6}
 & 0.10 & 75.63 & 0.5874 & 76.00 & 39.25 \\ 
\cline{2-6}
 & 1.00 & 90.79 & 0.5310 & 88.00 & 64.52 \\ 
\cline{2-6}
 & 10.00 & 99.10 & 0.4540 & 100.00 & 91.94 \\
\hline\hline
 \multicolumn{6}{c}{{\cellcolor[rgb]{0.773,0.773,0.773}}Xception backbone (different extractor)} \\ 
\hline
\multirow{4}{*}{LFW} & 0.00 & 91.90 & 0.4933 & 22.00 & 12.15 \\
\cline{2-6}
 & 0.10 & 97.76 & 0.4463 & 40.00 & 26.43 \\ 
\cline{2-6}
 & 1.00 & 99.41 & 0.4018 & 58.00 & 42.27 \\ 
\cline{2-6}
 & 10.00 & 99.90 & 0.3522 & 76.00 & 61.88 \\
\hline\hline
\multirow{4}{*}{\begin{tabular}[c]{@{}c@{}}color\\FERET\end{tabular}} & 0.00 & 99.64 & 0.5078 & 6.00 & 6.99 \\ 
\cline{2-6}
 & 0.10 & 99.82 & 0.4819 & 8.00 & 9.14 \\ 
\cline{2-6}
 & 1.00 & 100.00 & 0.4300 & 34.00 & 32.26 \\ 
\cline{2-6} 
 & 10.00 & 100.00 & 0.3738 & 66.00 & 61.29 \\
\hline\hline
\multicolumn{6}{c}{{\cellcolor[rgb]{0.773,0.773,0.773}}InceptionResNet backbone (different extractor)} \\ 
\hline
\multirow{4}{*}{LFW} & 0.00 & 79.77 & 0.5316 & 4.00 & 2.12 \\
\cline{2-6}
 & 0.10 & 98.12 & 0.4382 & 38.00 & 25.32 \\ 
\cline{2-6}
 & 1.00 & 99.42 & 0.3999 & 52.00 & 38.21 \\ 
\cline{2-6}
 & 10.00 & 99.86 & 0.3522 & 68.00 & 59.39 \\
\hline\hline
\multirow{4}{*}{\begin{tabular}[c]{@{}c@{}}color\\FERET\end{tabular}} & 0.00 & 99.46 & 0.528 & 4.00 & 2.69 \\ 
\cline{2-6}
 & 0.10 & 100.00 & 0.4655 & 22.00 & 13.44 \\ 
\cline{2-6}
 & 1.00 & 100.00 & 0.4239 & 46.00 & 39.78 \\ 
\cline{2-6}
 & 10.00 & 100.00 & 0.3712 & 76.00 & 67.74 \\
\hline
\end{tabular}
\end{table}

We first examine the attack performance in Sys T with the same feature extractor as Sys C, i.e., InsightFace with ResNet-50 backbone. As shown in Table \ref{tab.syst_res50}, both type-I and type-II attacks can achieve remarkable SARs under this setting. Even at 0\% FAR, the SAR can reach 86\% and 13.63\% for type-I and type-II attacks on LFW, respectively. The result from Table \ref{tab.syst_res50} suggest the proposed reconstruction scheme is effective in Sys T when using the same feature extractor as Sys C. 

Next, we evaluate the attack performance in Sys T with different feature extractors, including the Xception backbone-based InsightFace model and InceptionResNet backbone-based InsightFace model. As the results of the Xception and InceptionResNet backbones from Table \ref{tab.syst_res50} suggest, the proposed scheme can achieve good attack performance in both cases. Specifically, for the Xception-based extractor, 40\% and 26.43\% of the attempts can be successful under type-I and type-II attacks under 0.1\% FAR, respectively. For the InceptionResNet-based extractor, 38\% and 25.32\% of the attempts succeed in the type-I and type-II attacks under 0.1\% FAR on LFW, respectively. It is also worth noting that type-II attacks in both cases achieve comparable SARs, which validates the proposed reconstruction scheme's effectiveness. Furthermore, the results on colorFERET show similar promising performance, as indicated in Table \ref{tab.syst_res50}.

\begin{figure*}[t!]
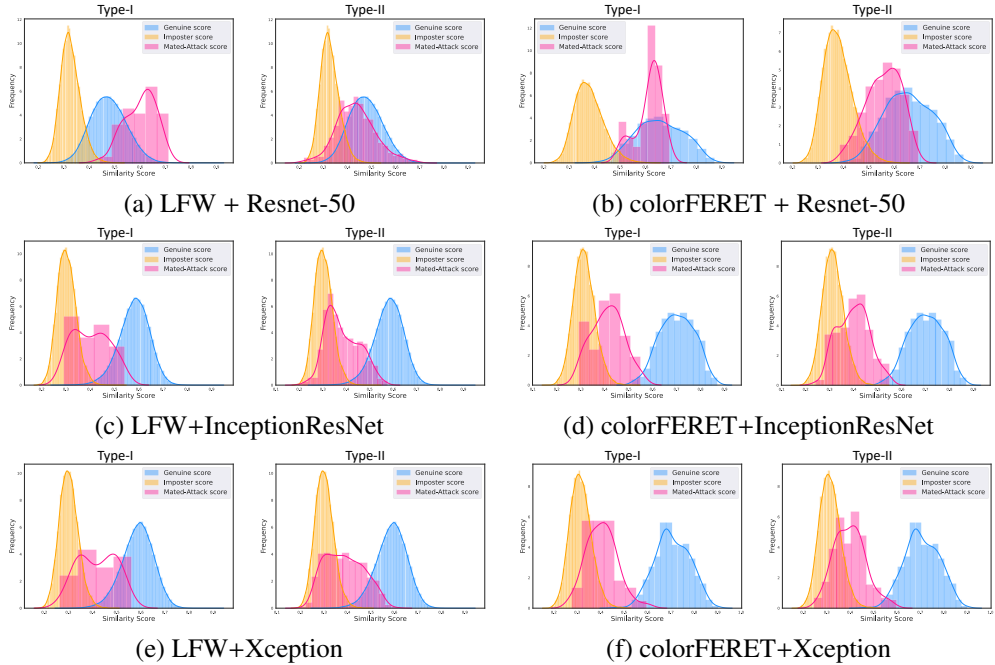

\centering
\imgStubPDFpage{0.49}{page=6,width=0.99\linewidth,trim=0cm 12.8cm 17cm 0cm, clip}{figures/f2f_diagram2.pdf}{(a) LFW + Resnet-50} 
\imgStubPDFpage{0.49}{page=6,width=0.99\linewidth,trim=17cm 12.8cm 0cm 0cm, clip}{figures/f2f_diagram2.pdf}{(b) colorFERET + Resnet-50} 
\imgStubPDFpage{0.49}{page=7,width=0.99\linewidth,trim=0cm 12.8cm 17cm 0cm, clip}{figures/f2f_diagram2.pdf}{(c) LFW+InceptionResNet} 
\imgStubPDFpage{0.49}{page=7,width=0.99\linewidth,trim=17cm 12.8cm 0cm 0cm, clip}{figures/f2f_diagram2.pdf}{(d) colorFERET+InceptionResNet} 
\imgStubPDFpage{0.49}{page=8,width=0.99\linewidth,trim=0cm 12.8cm 17cm 0cm, clip}{figures/f2f_diagram2.pdf}{(e) LFW+Xception} 
\imgStubPDFpage{0.49}{page=8,width=0.99\linewidth,trim=17cm 12.8cm 0cm 0cm, clip}{figures/f2f_diagram2.pdf}{(f) colorFERET+Xception} 
\caption{Distribution of the normalized similarity score of type-I and type-II attack on different datasets and different extractor backbones. (a) LFW and Resnet-50; (b) colorFERET and Resnet-50; (c) LFW and InceptionResNet; (d) colorFERET and InceptionResNet; (e) LFW and Xception; (f) colorFERET and Xception; 
\label{Figure:f2f_distribution}}
\end{figure*}

The normalized similarity score distributions of the genuine pairs, imposter pairs, and mated-attack pairs are shown in Figure \ref{Figure:f2f_distribution}. There is a gap between the distribution of the mated-attack scores generated from the reconstructed face images and the normal imposter score distribution. Specifically, when the attack is performed upon the same feature extractor, i.e., the ResNet-50, there is considerable overlap between the attack imposter score distribution and the genuine score distribution. On the other hand, an overlap exists between the attack imposter score distribution and genuine score distribution for type-II attacks under different feature extractors. The result from Figure \ref{Figure:f2f_distribution} indicates the effectiveness of our proposed method in terms of the similarity score distribution. Note that face recognition models with various backbones are different from the original InsightFace paper, hence the genuine and imposter has a overlap area.

\begin{figure*}[h]
\centering
\imgStubPDFpage{0.320}{width=0.99\linewidth}{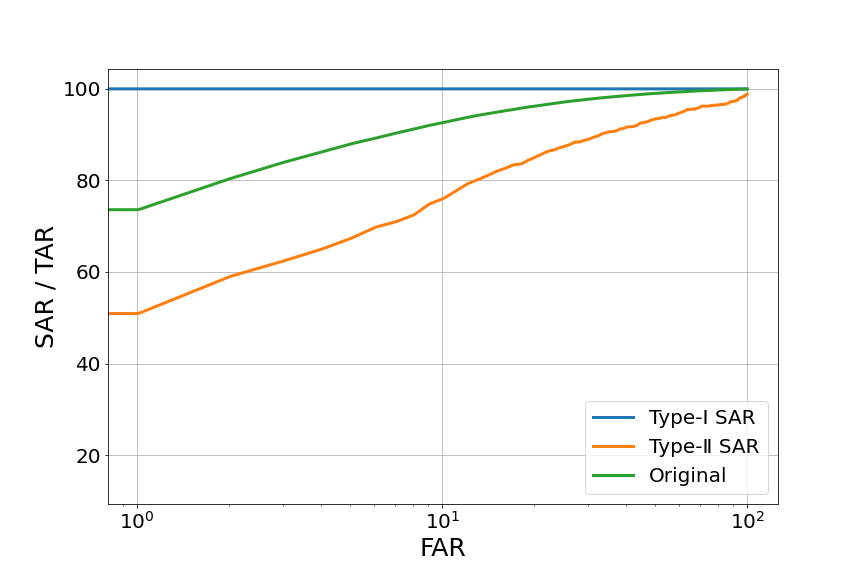}{(a) LFW+Resnet50 }
\imgStubPDFpage{0.32}{width=0.99\linewidth}{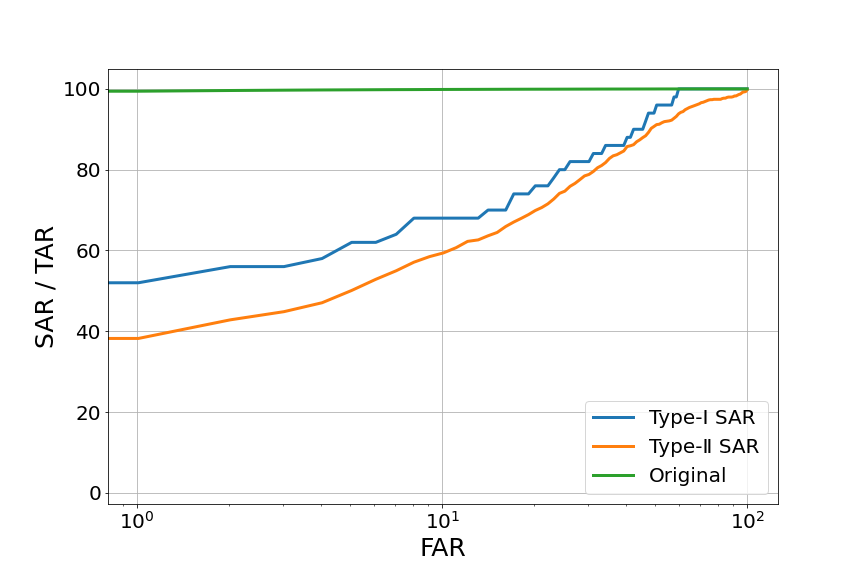}{(b) LFW+Inception }
\imgStubPDFpage{0.32}{width=0.99\linewidth}{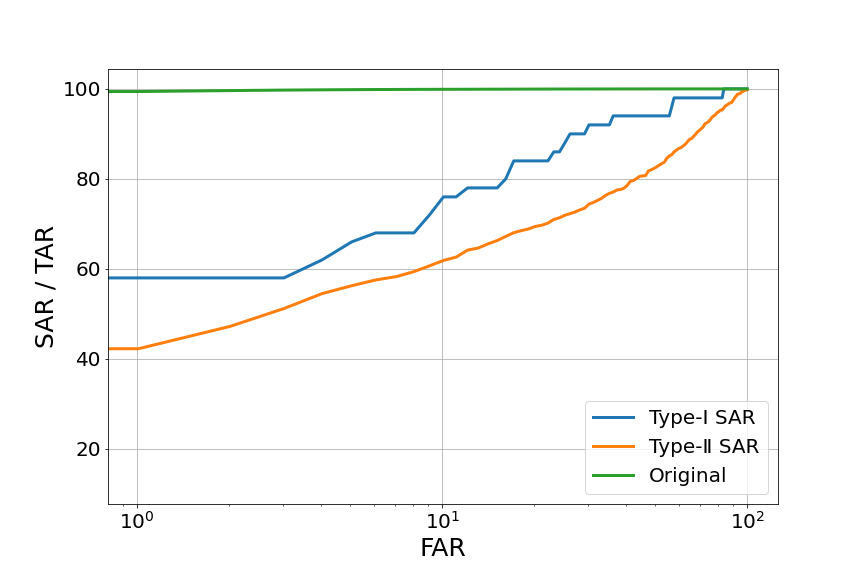}{(c) LFW+Xception }

\imgStubPDFpage{0.32}{width=0.99\linewidth}{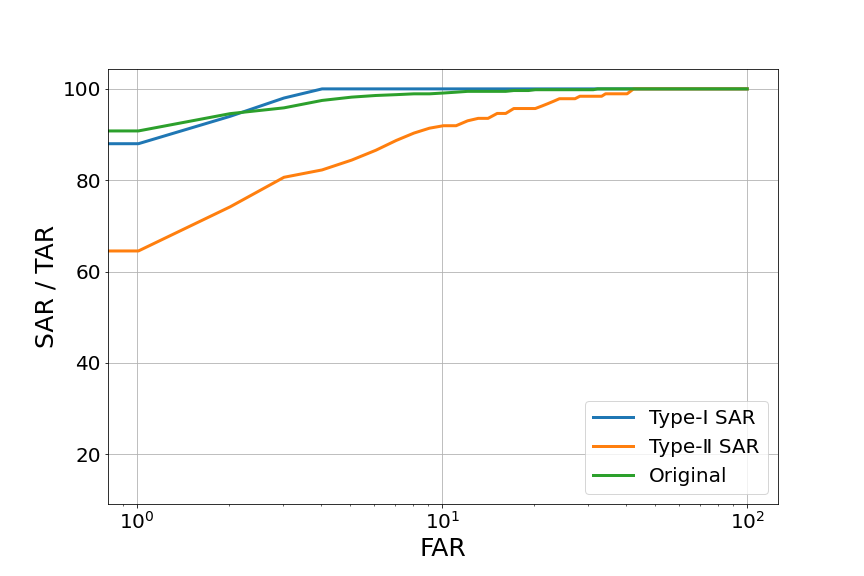}{(d) colorFERET+Resnet50 }
\imgStubPDFpage{0.32}{width=0.99\linewidth}{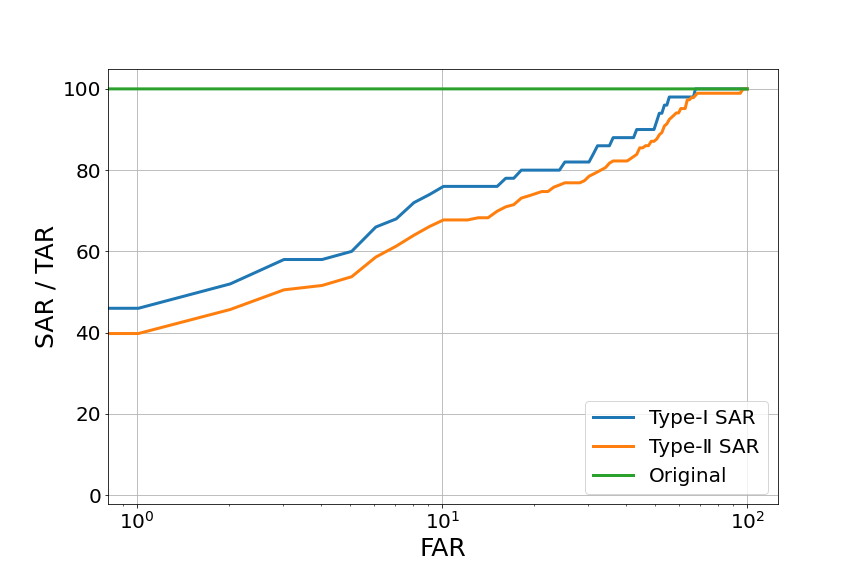}{(e) colorFERET+Inception}
\imgStubPDFpage{0.32}{width=0.99\linewidth}{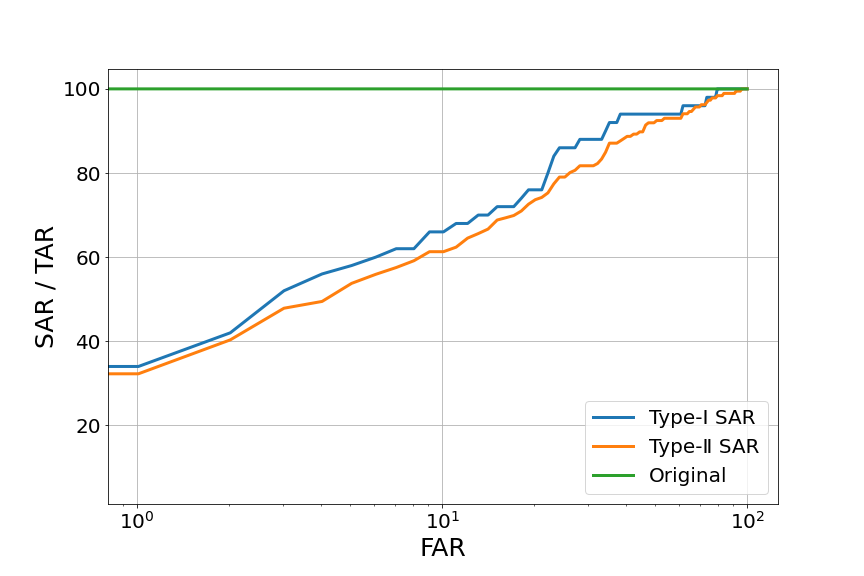}{(f) colorFERET+Xception}
\caption{ROC curves of type-I and type-II attack on LFW and colorFERET with various feature extractors in Sys T (FAR is logarithmic  scaled).\label{Figure:roc}}
\end{figure*}

The receiver operator characteristic (ROC) curves of
type-I and type-II attacks on LFW and colorFERET are shown in Figure \ref{Figure:roc}. The original ROC curve under a normal situation is also included in the graph. The results show that both type-I and type-II attacks can achieve remarkable performance in Sys T when using the same feature extractor with Sys C (see Figure \ref{Figure:roc}(a,d)). Meanwhile, the proposed scheme can also achieve a good attack performance when different feature extractors with Sys C are applied (see Figure \ref{Figure:roc}(b,c,e,f)).

A comparison with current state-of-the-art schemes can be found in Table \ref{table.sota}. We include the most recent closely-related state-of-the-art works, i.e., \cite{mai2018reconstruction,keller2021inverting,dong2021towards}, in our comparison. \cite{cole2017synthesizing} is not included as it is originally designed to generate frontal and neutral expression face images under white-box setting. \cite{zhmoginov2016inverting,razzhigaev2020black} are also not included due to the lack of attack performance data from the original paper. Compared with \cite{mai2018reconstruction}, face images can be produced in our pipeline without model training. Compared with \cite{dong2021towards}, the attack success rate has been boosted remarkably on the LFW dataset. Though \cite{mai2018reconstruction,keller2021inverting} can achieve better performance under Type-II attack setting, the generated images suffer from poor resolution quality.

\begin{table}
\centering
\caption{Comparison with the state-of-the-art methods\label{table.sota}}
\label{table.sota}
\resizebox{\linewidth}{!}{%
\begin{tabular}{l|l|c|c|c|c} 
\hline
\multicolumn{1}{c|}{Method} & \multicolumn{1}{c|}{Property} & \begin{tabular}[c]{@{}c@{}}Feature \\extractor\end{tabular} & Res. & \begin{tabular}[c]{@{}c@{}}Type-I attack\\on LFW \\@ 0.1\%FAR\end{tabular} & \begin{tabular}[c]{@{}c@{}}Type-II attack\\on LFW\\@ 0.1\%FAR\end{tabular} \\ 
\hline
Mai et al. \cite{mai2018reconstruction} & \begin{tabular}[c]{@{}l@{}}Neighborly de-convolutional~\\neural network (NbNet)~\\ trained on GAN-generated images\end{tabular} & Facenet & $160\times160$ & \begin{tabular}[c]{@{}c@{}}95.20\%\\(VGG-NbA-P)\end{tabular} & \begin{tabular}[c]{@{}c@{}}53.91\%\\(VGG-NbA-P)\end{tabular} \\ 
\hline
Keller et al. \cite{keller2021inverting} & \begin{tabular}[c]{@{}l@{}}Neighborly de-convolutional~\\neural network (NbNet)\end{tabular} & Facenet & $160\times160$ & \begin{tabular}[c]{@{}c@{}}95.16\%\\(Unprotected)\end{tabular} & \begin{tabular}[c]{@{}c@{}}65.05\%\\(Unprotected)\end{tabular} \\ 
\hline
Dong et al. \cite{dong2021towards} & \begin{tabular}[c]{@{}l@{}}Mapping between latent \\space and feature space\end{tabular} & InsightFace & \textbf{$1024\times1024$} & 1.42\% & 0.46\% \\ 
\hline
\textbf{Ours} & \begin{tabular}[c]{@{}l@{}}GAN's distribution modeling\\as a constraint of the~\\optimization task, solved by GA\end{tabular} & InsightFace & \textbf{$1024\times1024$} & \textbf{98.00\%} & 29.37\% \\
\hline
\end{tabular}
}
\end{table}

\subsection{Commercial-Off-The-Shelf liveness detection evaluation}
To further validate the advantage of the reconstructed face images, three COTS liveness detection cloud computing APIs from Asia and Europe areas\footnote{https://ai.baidu.com/tech/face/faceliveness; \\http://facebody.cn-shanghai.aliyuncs.com/?Action=DetectLivingFace;\\ https://skybiometry.com/demo/face-detect/ (threshold 50\%)} are used to evaluate the generated face images. The selected API depends solely upon a single RGB image for input. Different COTSs return different detection suggestions, but the detection results typically include the liveness scores and thresholds under various false reject rates. COTS 2 also produces suggestions for spoofing detection. COTS 3 produces liveness confidence for each image. Therefore, 150 reconstructed images from LFW are selected for evaluation. Furthermore, 150 random images were reconstructed from LFW, based on the methods \cite{mai2018reconstruction} and \cite{dong2021towards}, and are also selected for the same evaluation. The detailed results are shown in Table \ref{tab.liveness}.

\begin{table}[t!]
\centering
\caption{Liveness and spoofing detection by COTS.\label{tab.liveness}}
\begin{tabular}{l|c|l} 
\hline
Scheme & API & \multicolumn{1}{c}{Detection Details (150 images)} \\ 
\hline
\multirow{3}{*}{Dong et al. \cite{dong2021towards}} & \multicolumn{1}{l|}{COTS 1} & pass: 77, blocked: 17, need manual review: 56 \\
\cline{2-3}
 & \multicolumn{1}{l|}{COTS 2} & pass: 41 , block: 109, not spoofing: 3, spoofing: 147 \\
\cline{2-3}
 & \multicolumn{1}{l|}{COTS 3} & pass: 138, block: 12\\
\hline
\multirow{3}{*}{Mai et al. \cite{mai2018reconstruction}} & \multicolumn{1}{l|}{COTS 1} & pass:0, blocked:93, need manual review: 57 \\ 
\cline{2-3}
 & \multicolumn{1}{l|}{COTS 2} & pass: 71, block: 79, not spoofing: 150, spoofing:0 \\
\cline{2-3}
 & \multicolumn{1}{l|}{COTS 3} & pass: 0, block: 150\\
\hline
\multirow{3}{*}{\textbf{Ours}} & COTS 1 & pass: 133, blocked: 3, need manual review:14 \\ 
\cline{2-3}
 & COTS 2 & pass: 93, blocked: 57, is spoofing:2, not spoofing: 148 \\ 
\cline{2-3}
 & \multicolumn{1}{l|}{COTS 3} & pass: 147, block: 3\\
\hline
\end{tabular}
\end{table}

\subsection{Ablation Study}


There are three basic parameters of GA, i.e., population size, selection rate (crossover ratio), and mutation ratio. The population size determines how many individuals (the latent vectors) are in the population (in one generation). The selection rate determines how often crossover will occur, while the mutation ratio determines how often parts of an individual are mutated. 

\begin{figure*}[t!]
\centering
\imgStubPDFpage{0.32}{width=0.99\linewidth}{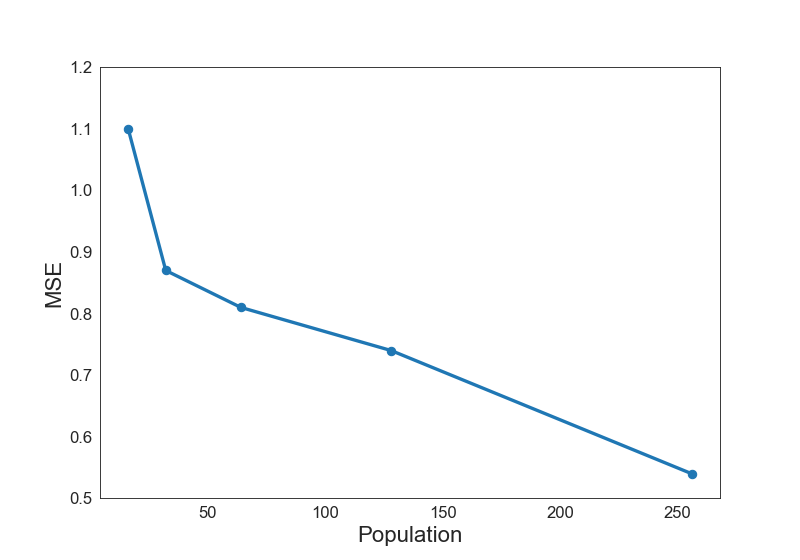}{(a) MSE vs population size}
\imgStubPDFpage{0.32}{width=0.99\linewidth}{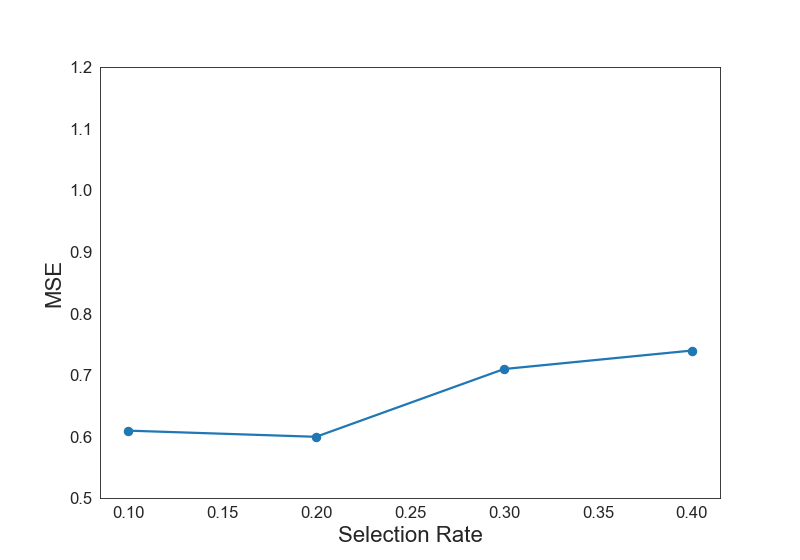}{(b) MSE vs crossover ratio}
\imgStubPDFpage{0.32}{width=0.99\linewidth}{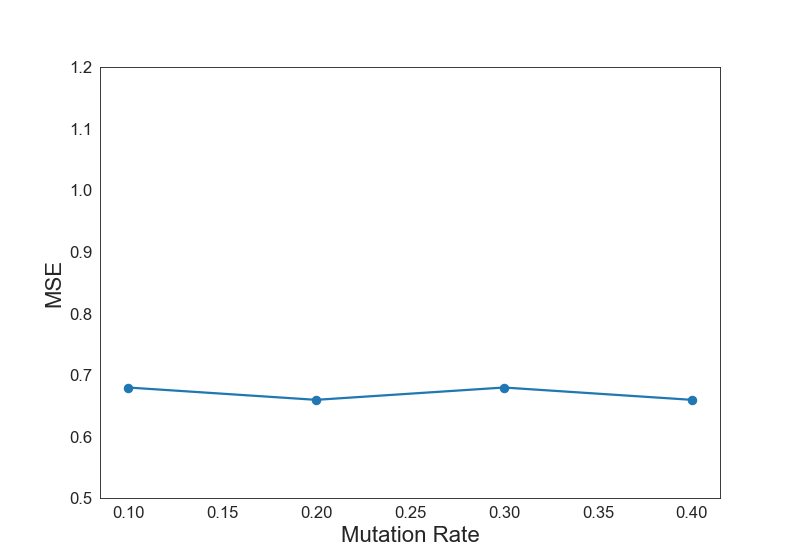}{(c) MSE vs mutation ratio}
\caption{ Ablation study on population size, crossover ratio, and mutation ratio. \label{Figure:ablation}}
\end{figure*}

To find the best population size, we set the crossover probability to 0.2 and the mutation ratio to 0.1. The fitness value in terms of mean square error (MSE) is evaluated at different ranges (16,32,64,128,256) of population size. The lower the MSE, the better performance is achieved. As shown in Figure \ref{Figure:ablation} (a), the MSE decreases while the population size increases. Generally, a smaller population size leads to smaller search space.
When the population size becomes large, GA requires more computational overhead and more time is needed. We notice that the 256 population size can perform well; meanwhile, the computation time and memory requirement remain acceptable.

To determine the best crossover ratio, we set the population size to 256 and the mutation ratio to 0.1. The fitness value is evaluated under various crossover ratio values (0.1,0.2,0.3,0.4). As shown in Figure \ref{Figure:ablation} (b), the best performance is achieved on 0.2. Similarly, the MSE under different mutation rates is shown in Figure \ref{Figure:ablation} (c); we can conclude that the impact of mutation rates with respect to the performance is not significant, so set the mutation ratio to 0.1.

\subsection{Failure cases}
\begin{figure*}[t]
 \centering
 \includegraphics[width=0.99\linewidth]{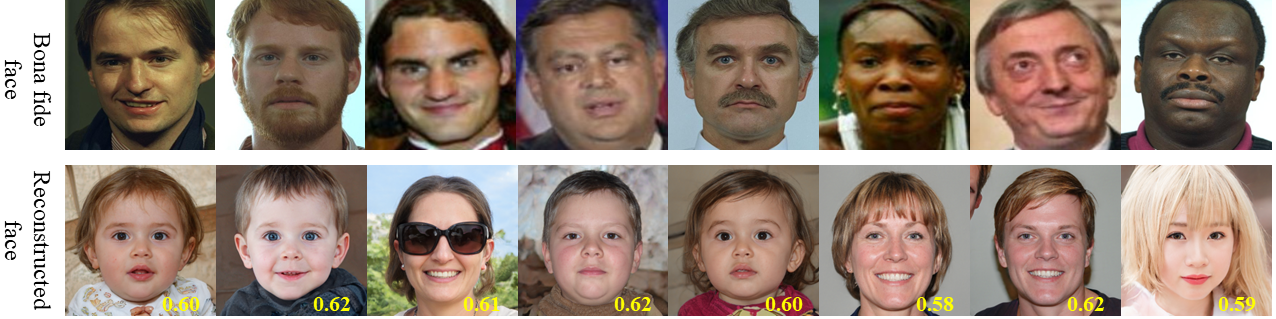}
 \caption{Failure cases.}
 \label{Figure:f2f_failurecase}
\end{figure*}

Several typical failure cases are shown in Figure \ref{Figure:f2f_failurecase}. In some failure cases, the fake face images are evolved easily into a baby. In other cases, the face images are evolved into female images. The failure of those cases is mainly the consequence of the generator's modeling bias, as the dataset used to train the StyleGAN may not be well balanced.

\section{Discussion and Conclusion}
In this paper, a method to generate face images from deep features has been presented. The reconstruction problem is formulated as a task to find the best latent vector of a pre-trained StyleGAN model, which can be utilized to generate a face image visually similar to the target face image. The generator is used as a human face constraint to the optimization task, and we propose to solve it with a genetic algorithm. Our experiment results prove that the generated face images show high similarity with the target face. We also utilize the generated face images to simulate type-I and type-II attacks on the LFW and colorFERET datasets. The quantitative results validate that the reconstructed face images can achieve a state-of-the-art attack success rate on the LFW dataset and comparable performance on colorFERET datasets. 

Three COTS liveness detection APIs are adopted to evaluate the generated face images. The results show that 88.6\% (62\%, 98\%) of generated images can pass the liveness detector COTS 1 (COTS 2, COTS3). This implies that the current COTS single photo-based liveness detection API in the market may not be safe enough. 

The forged face images in this work can shed light on current face recognition systems' security and privacy risks. The adversary could potentially exploit the generated face images to gain illegal access to the face recognition systems. Besides, using the GAN generator as an explicit distribution constraint proves to be effective and straightforward in our work; hence it can be potentially generalized to related constraint optimization tasks, e.g., inverting output to input.

It is worth pointing out that the proposed approach can also be applied to other privacy- and security-sensitive features such as text and video. To mitigate such risks, template protection schemes can be utilized to protect privacy- and security-sensitive data and better anti-spoofing detection are needed to avoid such risks.


\bibliographystyle{elsarticle-num}
\bibliography{egbib}

\end{document}